\documentclass[twocolumn,10pt]{asme2ej}

\usepackage{graphicx}
\usepackage{amsmath}
\usepackage{hyperref}   
\hypersetup{
	colorlinks=true,
	linkcolor=blue,
	citecolor=blue,
	urlcolor=blue,
}
\usepackage[square,numbers]{natbib}

\title{Design and Development of a Tracked Inspection Robot}

\author{\qquad Erika Sahari} \author{\qquad Weiyao Lai} \author{\qquad Alireza Pulles} \author{\qquad XiaoQi Guo} \author{Marc Bernhard}

\begin{document}

\maketitle    

This paper presents the examination of the clever Differential with three levels of opportunity. The is the principal differential with that interprets differential speed and force to its three results when the results are under fluctuated loads, however deciphers equivalent movement and force to its results when exposed to approach loads. The kinematics and elements of the are determined and are hypothetically investigated under three different burden cases. The movement of the under the three burden cases is additionally recreated and concentrated in. The benefits of alongside its current and potential applications are introduced.

\section{Introduction}

Pipelines are overwhelmingly utilized in ventures for the transportation of gases, oils and different liquids \cite{adriansyah2017Optimization}. They require successive assessment and support to forestall harm because of scaling and consumption. Because of their detachment, pipeline reviews are frequently convoluted and costly, for which mechanical investigation is a doable arrangement \cite{ravina2010low}. A wide assortment of drive-components have been investigated in the previous many years, for example, wheeled, screw, followed, pipe review measure, inchworm, enunciated and barely any others \cite{vadapalli2019modular,suryavanshi2020omnidirectional,vadapalli2021modular}. Be that as it may, the majority of them utilized various actuators and dynamic directing which expanded the control endeavors to guide and move inside the line, making off base confinement due slip while navigating in twists. By and large, three-followed in-pipe robots are demonstrated to be progressively more steady with better versatility \cite{ravina2016kit,choi2007pipe,kumar2021design,fujun2013modeling}. Our recently evolved robots \cite{vadapalli2019modular} and the Omnidirectional Tractable Three Module Robot \cite{suryavanshi2020omnidirectional}, both consolidated three-followed driving frameworks with dynamic differential speed to flawlessly arrange twists in pipe organizations. The paces of the three tracks of the robot were predefined in agreement to the twists in the line organization to decrease the slip and drag of its tracks. Chen Jun et al. \cite{chang2017development}, \cite{litopic}, \cite{vadapalli2021design}, \cite{vadapalli2021modular,9635853} are comparative three-module pipe climbers that works utilizing different drive-types (allude, contrasting their functionalities and our proposed robot). Deciding the paces of as far as possible the robot to explore twists just in directions for which the rates are preset \cite{vadapalli2019modular,suryavanshi2020omnidirectional}. Accordingly, the requirement for arranging as far as possible the robot's prosperity to just known conditions.

The referenced impediment can be tended to by utilizing a differential system which prepares the robot to work its tracks with differential speed inactively. One such differential, the multi-pivot differential stuff, was executed in the in-pipe robot \cite{deur2010modeling}. Consolidating a differential instrument empowered to lessen the slip and drag of its wheels extensively. Nonetheless, the stuff plan utilized in the differential is with the end goal that the system favors one of its results $($output$)$ over the other two results $($output and yield$)$ as found in the schematic \cite{9517351}. As an outcome, when the robot navigates in pipes, one of the tracks moves quicker than the other two causing slip or drag in a couple of directions of the robot \cite{9517351}. This impediment is unfolded on the grounds that every one of the three results of the differential don't have comparable elements with the information. Answers for three-yield differentials  were likewise introduced by S. Kota and S. Bidare and Diego Ospina and Alejandro Ramirez-Serrano \cite{armstrong1992error}. The differentials they proposed have schematics like and display comparable impediments as the multi-pivot differential stuff. Because of the constraints present in the as of now existing arrangements, there exists the requirement for a three-module robot that crosses in-pipe with next to no slip or drag.

We propose , third in the series of our three-followed pipe climbing robots. Our answer involves consolidating a differential component to take out the slip and drag made in robot's tracks due the evolving cross-segment of the line while arranging twists. Resultantly, we make two key commitments. First and foremost, plan of the   Differential  \cite{vadapalli2021modular,9635853,saha2022pipe}.  is the principal differential with every one of the three results sharing an identical kinematics relations to the contribution, as displayed in the. Accordingly,  has the clever capacity to decipher equivalent rates and force to every one of results are under equivalent burdens, as hypothetically addressed in conditions Furthermore, consolidating the  improves the robot's capacity to cross line networks with next to no slip or drag in any robot-direction.

The robot is intended to cross inside lines of measurement mm to mm without requiring any dynamic control. The paper talks about exhaustively the plan of the robot and the clever component it consolidates. The kinematics of component and the robot is planned and the robot's line exploring capacities are approved through tests.

\section{Development of the Robot}

The  contains three tracks that are controlled by a solitary engine by means of the , as seen in. Tracks are housed on isolated modules which are fixed separated from one another on a nonagon-molded focus case, as displayed in. The modules are associated with the body through wall-clip system which push the tracks against the internal mass of the line and give footing, as represented in. The robot measures long and in distance across.

The tracks comprises of hauls between associated by chains that is pivoted by the sprockets. The plan of the tracks depend on the two finishes of the sprocket for creating the tractive power. In this way, when the robot promptly enter the curves from straight line segments, the driving sprocket and the determined sprocket in the track will have a general distinction in their precise speeds. To determine this, each track is given a little leeway. The negligible leeway helps in changing the strain of the track when there is a distinction in the precise speed between the two closures of the sprocket in a similar module. Accordingly, the tracks are obliged close to the curves and the tractive power is constantly kept up with between the line and the track during twists and straight lines. The robot's capacity to work in pipes is predominantly ascribed to two components, the wall-clip system and the 

\begin{figure}[ht!]
\centering
\includegraphics[width=3.1in]{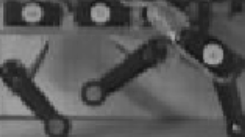}
\caption{\footnotesize Tracked Type
Pipeline Inspection Robot}
\label{1}
\end{figure}

The inward wall-cinch component acts freely on every module as a consistent framework. It contains straight springs, linkages and plug as displayed in a. It is situated between the suspension and the modules to give outspread consistence to the robot. Every module is spring stacked and associated with skeleton by four linkages. The modules have the arrangement to slide along the linkages. Every linkage houses a direct spring which is pre-stacked and pushes the module radially outwards. The plugs guarantee that the modules are not stretched past the boundary. The consistence empowers the robot to inactively shift its measurement by from to keep up with foothold and adjust to different circumstances the robot could insight inside a line. What's more, it permits every modules to pack lopsidedly to cross over impediments, as outlined in b.

The  is a clever component of the . The differential is fitted inside the body and is associated with the three tracks through angle sprocket game plan, displayed in. Utilizing the differential to drive the robot possibly dispenses with the slip and drag of its tracks. This significantly decreases the weights on the robot and hence, giving a smoother movement to the robot. The  contains a solitary info, three two-yield  differentials, three   differentials and three results, as displayed in. The differential's feedback is situated at its middle and the three two-yield differentials are organized around the contribution with a point of between them. The  differentials are fitted evenly between the two-yield differentials. The single result of every one of the three  differentials structure the three results of the , as seen in.

\begin{figure}[ht!]
\centering
\includegraphics[width=1.7in]{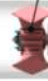}
\caption{\footnotesize Sectional View}
\label{1}
\end{figure}

The information from the worm gear all the while gives movement to the three two-yield differentials which further makes an interpretation of the movement to its adjoining  differentials relying upon the heap their side pinion wheels experience. The make an interpretation of differential speed to its adjoining on the off chance that its two side pinion wheel work under various burdens. The movement got by the two side pinion wheels of the individual  differentials is meant the three results. At the point when the two side pinion wheels of a  differential get various paces, it makes an interpretation of the differential speed to a solitary result. The six differentials and work together to decipher movement from the information to the three results.

Novelty of The three results of the  have comparable contribution to yield elements which can be seen in its schematic in. Moreover, the results share similar elements with one another. Thus, the adjustment of burdens for one of the results will similarly affect the other two results. Consequently, the  accomplishes the clever aftereffect of working its three results with differential speed when they are under shifted loads and with equivalent paces when the results are exposed to identical burdens, as hypothetically determined in the situations. For example, when the external module works at an alternate speed and the other two internal modules are under identical burdens in the curves, then, at that point, both the inward modules work with equivalent precise speeds and equivalent forces. This is one more clever outcome acknowledged by utilizing the  in the line climber. Attributable to these outcomes, the  is the initial three-yield differential whose working is comparable to that of the conventional two-yield  differential.

This  is explicitly intended to be utilized inside the . It furnishes the robot with differential speed inside a line twist so the track venturing to every part of the more extended distance turns quicker than the track venturing to every part of the more limited distance, yet while moving inside a straight line segment the three tracks pivot with identical rates.

\section{Design Specifications}

This segment presents the rakish speed and force connection between the results and the info. The kinematics and the elements of the  are inferred by the method for the bond chart displaying strategy \cite{vadapalli2021design}, showing the clever capacity of the instrument to display identical result to include precise speed and force.

By comparing the contribution to the side stuff condition and the result to the side cog wheels connection, we get the contribution to yield connection for the precise speed, allude \cite{vadapalli2021design}.

where is the stuff proportion of the contribution to the ring gears, while is the stuff proportion of the ring gears to the results. are the rakish speed and force of the info. Additionally, result to the info force relations are acquired Equal paces and force Equations show that every one of the three results of the differential offer identical rakish speed and force relations with the information. Since the side cog wheels are indistinguishable, they display equivalent idleness. At the point when each of the three results are under equivalent burdens restive force, where goes from subbing these relations in outline the clever capacity of the  to decipher equivalent precise speeds and force to all three results are unconstrained or under equivalent burdens. At the point when any of the results is obliged because of a heap, then the other two results act as per the variety of because of. The precise speeds of the three results decipher as direct speeds of the three tracks where is the width of the tracks. Shows the planar portrayal of the robot's route inside a line twist of point. The focal point of the line is checked indicates the range of curve of the line twist about the pivot are the resources of the three tracks of the robot with the internal mass of the line, with being the span of the line. The robot's middle matches the line's middle are the opposite distance between the tracks of the robot and the hub. Pipe-twists are typically intended to have a consistent span of shape. Subsequently, as the robot explores inside a line twist, the ways followed by its three tracks have a uniform curve, following which the distances stay steady. The differential in the robot prompts the track that is farther away from to travel quicker than the track that is nearer to. The three tracks of the robot are separated from one another. is the point subtended between the internal track of the robot to its sweep of bend of the line. It lies between the internal track of the robot is the symmetrical projection of the point on the hub. In the point among since the points. The lines interfacing are equal and consequently, the points are equivalent. Hence, the range of ebb and flow of the way followed by the track at Using the qualities span of the line, sweep of curve of the line twist, the velocities of the three tracks of the robot in any direction inside any line twist of point can be determined. The speed of the robot stays consistent in all directions and is generally equivalent to the normal of the rates of the three tracks. From the distance went by the robot and the information speed, the time taken displayed in to arrange the line is determined. The planned direct rates for each tracks at various directions were utilized to achieve the hypothetical distance.

\begin{equation}
(M^{'} + \mu M^{'}) \tan{\arcsin{(M^{'}k^{'})}} = H
\label{2}
\end{equation}

A model of the , as displayed in, is worked to additionally approve the robot's capacity to navigate pipe-organizations. The differential of the model is controlled by a engine which produces ostensible force. The robot is controlled remotely through Optical encoders are introduced on the robot to gauge the paces of the three tracks. The encoders' sensors record a perusing for each which is taken as the least count mistake for the examination. The robot is tried inside a line network worked to the components of the mimicked climate, developed as displayed.

\begin{equation}
K_s = \frac{\tan{\mu N}}{12}
\label{3}
\end{equation}

\begin{figure}[ht!]
\centering
\includegraphics[width=2.3in]{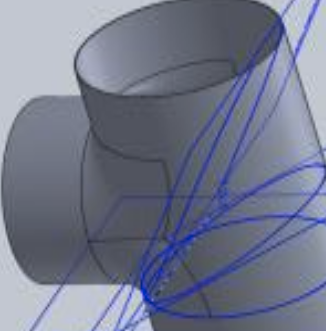}
\caption{\footnotesize Bend 1}
\label{asym(1)}
\end{figure}

\section{Experimentation}

The paces of the three tracks and the robot in cases are as displayed in. The robot's route in various segments of the line with the robot's direction being is displayed in Figure~\ref{1}. While navigating in vertical line and even line, the noticed mean direct velocities of the three tracks and the robot are. Construing from Fig.~\ref{2}, the paces of the three tracks and the robot in the line straights are roughly equivalent with a blunder of \cite{vadapalli2021modular,9635853,saha2022pipe}. While arranging the curve the mean direct speed of the track, ventures to every part of the briefest distance than the mean straight velocities of track. At direction displayed in Figure~\ref{3}(a), we can find that the track, which ventures to every part of the longest distance, moves the quickest among the three tracks. The robot takes seconds to explore the line organization. The absolute distance went by the three tracks and the robot in the direction separately, as displayed.

\begin{figure}[ht!]
\centering
\includegraphics[width=2.2in]{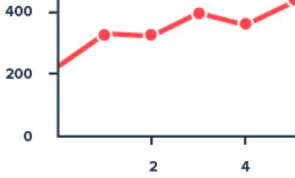}
\caption{\footnotesize Bend 2}
\label{asym(1)}
\end{figure}

Navigation without slip and drag, the distances recorded tentatively match the hypothetical outcomes as found in, with the most extreme blunder happening at track, which adds up to a rate mistake of. The recorded blunder distance determined hypothetically deducted from the distance recorded tentatively, for track is inside the least count mistake assessed for the trial, as displayed in Table II. The kept blunders somewhere far off went for the three tracks and the robot in every one of the three cases are inside the assessed least count mistake with the most noteworthy deviation happening at track direction, as represented. These outcomes support our recommendation that the robot crosses with next to no slip or drag. Besides, it is tentatively seen that the paces of the three tracks and the robot, in every one of the three directions in show a severe similarity with hypothetical outcomes as seen.

This further states the robots capacity to explore without slip and drag in any direction of the robot. The  accomplishes this original outcome since every one of the three results from the  component have comparable precise speed and force conveyance to one another and to the information. The three tracks work at practically equivalent speeds while moving in pipe-straights. In the line twists, the track speeds adjust as per the range of curve of the line at any embedded direction of the robot \cite{9197210,vadapallig}. The moment distinctions in the speeds can be credited to the resistances present in the plan and the trial and error technique. The line climber is likewise effectively tried in other irregular directions.

\section{Conclusion}
The integrating the original   Differential is introduced. The differential is planned with the end goal that its working skill is undifferentiated from the customary two-yield  differential. The differential empowers the robot to arrange pipe-twists without the requirement for any dynamic control and extra control endeavors. The robot is tried in a perplexing line network in numerous directions and the outcomes check the robot's capacity to explore different line segments in all directions with next to no slip or drag because of the evolving cross-part of the line. Keeping away from slip and drag extensively diminishes the burdens experienced by the robot and gives a smoother movement to the line climber.


\bibliographystyle{asmems4}

\bibliography{asme2e}

\end{document}